\documentclass{article}

\usepackage{arxiv}

\usepackage[utf8]{inputenc} 
\usepackage[T1]{fontenc}    
\usepackage{hyperref}       
\usepackage{url}            
\usepackage{booktabs}       
\usepackage{amsfonts}       
\usepackage{nicefrac}       
\usepackage{microtype}      
\usepackage{lipsum}
\usepackage{graphicx}
\usepackage{algorithm,algpseudocode}

\graphicspath{ {./images/} }

\title{GMFIM: A Generative Mask-guided Facial Image Manipulation Model for Privacy Preservation}

\author{
Mohammad Hossein Khojaste \\
  School of Computer Engineering\\
  Iran University of Science and Technology\\
  Tehran, Iran \\
  \texttt{m\_khojaste@comp.iust.ac.ir} \\
   \And
 Nastaran Moradzadeh Farid \\
  Computer Engineering Department\\
  Amirkabir University of Technology\\
  Tehran, Iran \\
  \texttt{nmoradzadehf@aut.ac.ir} \\
  \And
 Ahmad Nickabadi \\
  Computer Engineering Department\\
  Amirkabir University of Technology\\
  Tehran, Iran \\
  \texttt{nickabadi@aut.ac.ir} \\
}

\begin{document}
\maketitle
\begin{abstract}
The use of social media websites and applications has become very popular and people share their photos on these networks. Automatic recognition and tagging of people’s photos on these networks has raised privacy preservation issues and users seek methods for hiding their identities from these algorithms. Blurring or blacking the face area, adding physical adversarial patches to the face, and adding adversarial masks are some proposed methods for this purpose. However, these methods mainly suffer from two main problems: the output image does not look like the input image, or the identity is not effectively concealed from automatic face recognition (AFR) methods. \par
Generative adversarial networks (GANs) are shown to be very powerful in generating face images in high diversity and also in editing face images. In this paper, we propose a Generative Mask-guided Face Image Manipulation (GMFIM) model based on GANs to apply imperceptible editing to the input face image to preserve the privacy of the person in the image. Our model consists of three main components: a) the face mask module to cut the face area out of the input image and omit the background, b) the GAN-based optimization module for manipulating the face image and hiding the identity and, c) the merge module for combining the background of the input image and the manipulated de-identified face image. Different criteria are considered in the loss function of the optimization step to produce high-quality images that are as similar as possible to the input image while they cannot be recognized by AFR systems. The results of the experiments on different datasets show that our model can achieve better performance against automated face recognition systems in comparison to the state-of-the-art methods and it catches a higher attack success rate in most experiments from a total of 18. Moreover, the generated images of our proposed model have the highest quality and are more pleasing to human eyes. 
\end{abstract}


\section{Introduction}
Nowadays, social networks have become an essential part of the people's life all over the world. People spend a lot of time in the related websites and applications and share lots of personal photos and videos on them every day. For example, Facebook as the champion of social media platforms has over 2.74 billion active users, and Instagram, the most popular photo-sharing app has 1.221 billion active users \cite{karl2021}. On Instagram, 95 million photos and videos are shared per day \cite{lister2021}. However, privacy issues are now a common concern among the users of these networks. For example, due to the growing societal concerns about the use of automatic face recognition (AFR) systems for identifying, processing and tagging people on images and videos, some social networks have decided to remove this technology from their apps. However, many people still seek solutions for hiding their identities from AFR systems. A straightforward solution is to de-identify images before posting them on the social network by editing the face area. \par

There are two main requirements for concealing the identity of a face image by editing. First, the manipulated image should be as similar as possible to the input image so that a human observer can still identify the target person from the generated face image. Also, the editing process should not add any noticeable artifacts to the initial image. Second, the generated image should not be identifiable as the target person by an AFR system. Unfortunately, most of the previous methods proposed for this purpose fail to meet both requirements simultaneously.  On one hand, due to the current advances in modern face recognition systems, it is very difficult to hide the identity of a face image from these systems. On the other hand, some models that can produce an acceptable attack success rate against these systems, have low-quality images or noticeable changes and hence not suitable for publishing on social media. The first idea was to blur or darken the face area of the image \cite{yang2021study}. While this method has a high success rate in hiding the identity, its main drawback is that the resulted images do not meet the first requirement and cannot be shared with others. The second category of methods replaces the faces of the input image with different faces so that the image seems normal to the human eyes but the identities are altered \cite{sun2018natural, li2019anonymousnet}-suitable for applications like broadcasting a street view on the web not sharing your personal photos with your friends. The goal of the last group of methods is to edit the face regions of the input image in such a way that the faces are still similar to the original ones but not recognizable by the AFR tools \cite{deb2020advfaces, oh2017adversarial, wu2018privacy, shan2020fawkes}. The two main challenges of these models is the low success rate against face recognition systems and the undesired visual effects added to the original image through the modification process. \par

Generative Adversarial Networks (GANs) are a deep learning-based approach to generative modeling that consists of two main components: a generator and a discriminator \cite{goodfellow2014generative}. These models have been successfully used in various image applications including image generation, image manipulation, and image-to-image translation. In the past few years, many GAN-based models have been proposed for face image generation and manipulation. In this paper, we propose a generative mask-guided face image manipulation (GMFIM) model that is based on GANs and can produce high-quality images without any noticeable distortion and is very effective against face recognition systems. Our model consists of three main components: a) a face mask module to remove the background and extract the face area of the input image as the target region of the de-identification process, b) a GAN-based optimization module which is the key component of our model and formulate the de-identification task as an optimization problem with two loss functions, and c) a merge module that employs a Fourier transform-based method \cite{szeliski2010computer} for combining the background from the input image to the output of the optimization module to give the final image. To evaluate our proposed model, we have used two different face datasets and the de-identification success rate against three well-known face recognition systems in 18 experiments are compare with two other state-of-the-art identity hiding models. In most of the experiments, GMFIM achieves better attack success rate while the quality of its generated images is better than those of the two other methods. \par

The rest of the paper is organized as follows. In Section \ref{sec:rel_works}, we present related works on de-identification methods and generative adversarial networks. Section \ref{sec:proposed_model} describes the proposed model in detail. In Section \ref{sec:experiments}, the results of our model are analyzed and compared with those of the state-of-the-art models. \par

\section{Related works}
\label{sec:rel_works}
In this section, we first review different de-identification methods and then briefly explain the Generative Adversarial Networks and some of their applications as our proposed de-identification model is based on these networks. The identity hiding methods studied in this section are categorized in three main groups. The first group consists of simple techniques such as blurring or blacking the facial regions of the input image. The second category is composed of methods that replace the faces of the input image with new faces with completely different identities using the so-called face-swap techniques. The third category of the de-identification methods manipulate each face image in such a way that the edited face looks like the original one by human observer but not recognizable by automatic face recognition systems. \par

\subsection{De-identification methods}
There are various methods for concealing the identity of the people who appear in a photo from intelligent algorithms. The very first attempts in this field were simple techniques for detecting and blurring or blacking the face regions in an image. The first drawback of these techniques is that the resulted images are not visually appealing to human eyes and cannot be shared on social networks. The second drawback is that the obfuscation methods like blurring and mosaicking cannot effectively protect the identity. In \cite{mcpherson2016defeating}, a deep neural network is trained that successfully identifies the faces which have been hided with the aforementioned techniques. However, recently these models have been utilized for preserving the privacy of the people in public image datasets \cite{orekondy2018connecting, yang2021study}.\par

The second category of de-identification methods unlike the first category changes the faces in an image in such a way that the image still appears normal to human eyes but the faces in the output image are quite different from those of the input image. For example, Sun et al. propose a model which firstly detects or generates the facial landmarks of the input image and secondly, takes the blackhead image and the set of landmarks as input and generates the inpainted image \cite{sun2018natural} which is a new face with the same facial landmarks and background. AnonymousNet  \cite{li2019anonymousnet} is another model of this category which is a four-stage framework that uses a deep convolutional neural network to sequentially  extract face semantics, alter the values of some facial attributes, generate a new face image employing a face generator, and finally perturb the generated image by adding adversarial noise. Also, Pautov et al. propose a different idea which can be considered as a physical attack to face recognition systems \cite{pautov2019adversarial}. In this method, an adversarial patch is designed, printed and then added to the face of the target person. The face of this person along with the attached adversarial sticker is photographed and the photo is then fed into face recognition systems to measure the success rate of the proposed framework. \par

As stated in \cite{deb2020advfaces}, one of the most important requirements of a face de-identification method is that the produced image be perceptually realistic-enough so that a human observer will recognize it as a legitimate face image compared to the set of input images. On the other hand, some applications call for the preservation of the identity for the human observers and hiding the identity from the automatic face recognition systems. The methods of the third de-identification category try to simultaneously address all of these requirements. For example, Oh et al. \cite{oh2017adversarial} propose a general game-theoretical framework for this purpose in which a user is playing against a recognizer; the user changes the input image based on a strategy (an adversarial image perturbation) for concealing the identity and the recognizer processes the generated image based on a different strategy for neutralizing the changes made by the user. The optimal strategy of the user is then used for identity protection. 
Wu et al. \cite{wu2018privacy} propose a GAN-based framework, called PP-GAN, in which the generator part is trained for transforming the input face image to a similar image with hided identity. In this model, first, a verification model is trained to determine whether two face images belong to the same person or not. They also use the pixel level structure similarity (SSIM) to make the generated image as similar as possible to the input image. The loss of the verification model and the SSIM loss are used to train the generator, and the final generator is used as the model for de-identify the input images.
In a more recent work, Deb at al. proposed AdvFaces \cite{deb2020advfaces} which is based on creating image-dependent adversarial masks for hiding identity. AdvFaces is also a GAN-based model consisting of a generator, a discriminator, and a face matcher; the generator creates an adversarial additive mask for the input image, the face matcher is used to remove the identity from the generated image, and the discriminator guides the model to generate perceptually realistic images. Finally, Shan et al. have developed Fawkes \cite{shan2020fawkes} for creating perturbations that added to a face image, conceal the identity from the intelligent algorithms. Fawkes seeks to find minimal perturbations that drastically shift the feature vector of the generated image in the feature space. \par

\subsection{Generative Adversarial Networks (GANs)}
Generative Adversarial Networks, initially proposed in \cite{goodfellow2014generative}, are a class of machine learning frameworks consisting of two main components: the generator and the discriminator.  The generator G captures the data distribution and the discriminator estimates the probability that a sample come from the real training data rather than G.  The training goal of G is to maximize the probability of D making a mistake. This framework corresponds to a two-player minimax game. \par

GANs have been used in a wide variety of image applications such as image in-painting \cite{dolhansky2018eye, demir2018patch}, image manipulation \cite{brock2016neural, he2019attgan}, object detection \cite{bai2018sod, prakash2019gan}, 3D image synthesis \cite{wu2016learning, ongun2018paired}, and image-to-image translation \cite{zhu2017unpaired, choi2018stargan}. \par

One of the most important applications of GANs is to generate high-quality images. For example, PGGAN \cite{karras2017progressive} is a high quality image generator whose main idea is progressive growing of the generator and discriminator. So, it starts with a low resolution image and increases the resolution by adding new layers so that the generated images have more fine details as the training progresses. As stated in \cite{jabbar2020survey}, despite the fact that PGGAN generates high-quality images, its ability to control specific features of the generated image is limited. StyleGAN is an upgraded version of PGGAN that modifies the architecture of the generator network and uses a mapping network to convert the input latent noise to the intermediate latent style vector. Adaptive Instance Normalization (AdaIN) is used to create styles and manage the layers of the synthesis network. In \cite{karras2020analyzing}, several characteristic artifacts of StyleGAN are analyzed, and several model architectures and training process adjustments are offered to overcome these issues. \par

\begin{figure*}[t]
\centering
\includegraphics[width=0.75\textwidth,height=14cm]{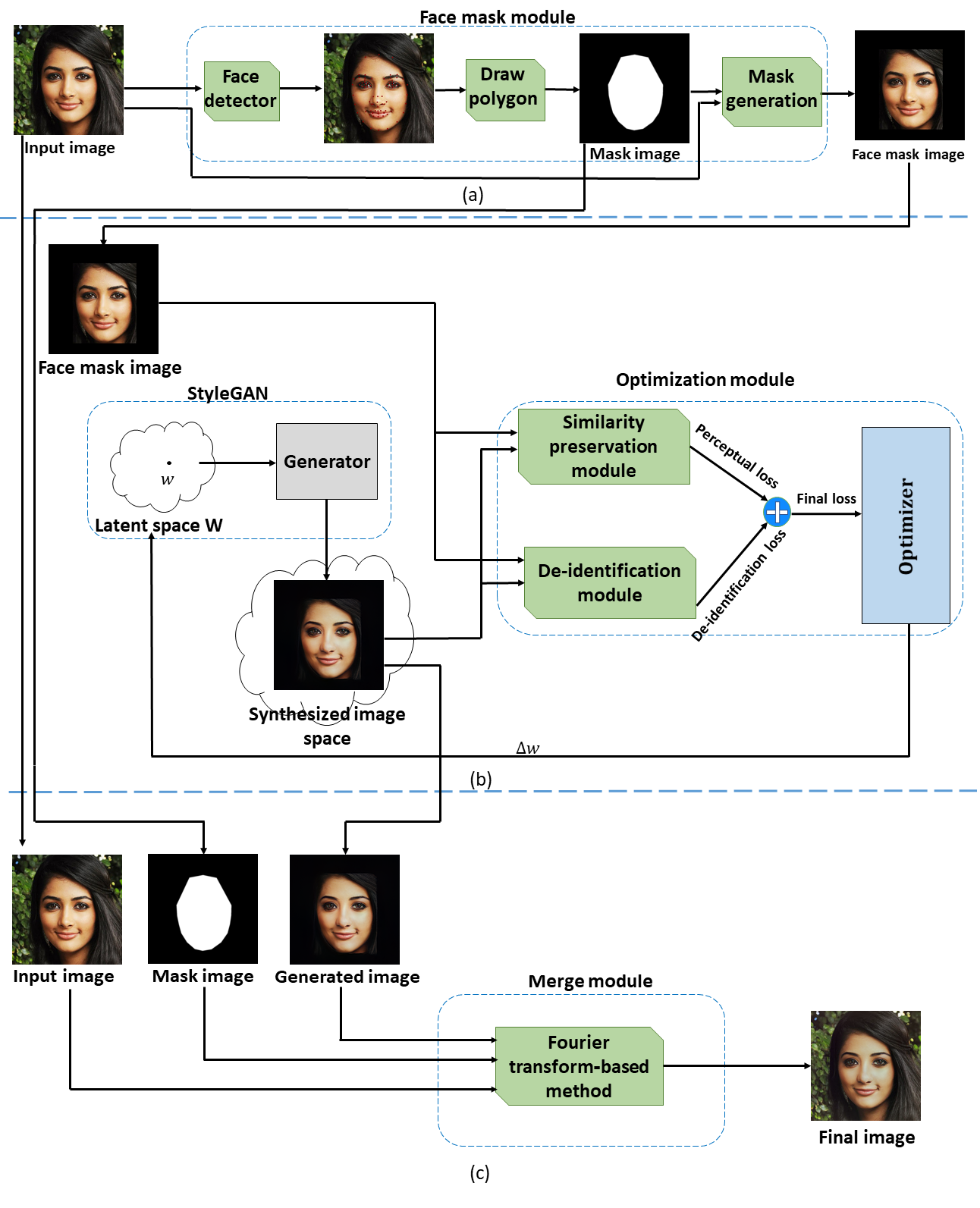}
\caption{The proposed GMFIM model. (a) The face mask module is used to extract the face part of the image to help the optimization module to focus only on the face part. In this module, at first, facial landmarks are extracted, and then face polygon is detected based on these landmarks. Finally, the rectangular area of the face is extracted from the input image and sent to the optimization step. (b) The optimization module performs an iterative optimization process to generate the de-identified image based on a face generator and two loss functions. (c) The merge module adds the original image background to the de-identified image generated by the optimization module using a Fourier transform-based method.}
\label{block_diag}
\end{figure*}

\section{The Proposed Model}
\label{sec:proposed_model}
In this section, we propose our de-identification model called GMFIM. As stated before, given an input image, the goal of this model is to create a face image that looks like the input image while automatic face recognition systems cannot match the synthesized image to the true identity. Our proposed model encompasses three main components: a) the face mask module, b) the optimization module, and c) the merge module. Figure \ref{block_diag} outlines the proposed model. The three components of the model are discussed in the following. \par

\subsection{Face mask module}
As our goal is to de-identify the input image, it is necessary and sufficient just to alter the face part of the image. Therefore, in the first step of the propose model, the face part of the input image is extracted and the background is omitted. This will help the optimization module to focus only on the face part of the image for both reconstruction and de-identification goals. \par

The proposed module takes an input image and produces an image with a black background. As shown in Figure \ref{block_diag} (a), firstly, the facial landmarks are extracted from the input image with the help of the face and landmark detector algorithm proposed in \cite{milborrow2014active}. Then, a rectangle box is formed around the face using the set of extracted landmarks. The rest of the image, regarded as background, is then removed (filled with black color).\par

\subsection{Optimization module}
The core component of GMFIM is the optimization module in which the de-identification task is formulated as an optimization problem and an iterative gradient descent algorithm is adopted to find the best solutions. \par

As illustrated in Figure \ref{block_diag} (b), the optimization module is based on a face generator that synthesizes realistic face images from random latent vectors. Here, the goal is to find a latent vector that is mapped to a face image with two required characteristics:  1) it perceptually looks like the face image extracted in the first step, and 2) the automatic face recognition systems cannot match it to the true subject. To achieve these goals we have defined two loss functions described in the following. The final output of this step is the face image that best minimizes these losses. \par

Regarding the face generator, we have utilized the pre-trained StyleGAN2 \cite{karras2020analyzing} for this part of our proposed model. However, our proposed model is a general framework and any other face generator can be used in this section without any changes in the model. \par

\textbf{Perceptual loss:} As stated before, the output image of this step should be perceptually realistic and similar to the input image but not exactly identical to it. Traditionally, the low-level similarity between two images has been measured in pixel space, but in recent years, this has changed \cite{abdal2019image2stylegan}. Gatys et al. observed \cite{gatys2015texture, gatys2015neural} that the learned filters of the VGG image classification model \cite{liu2015very} are very good general-purpose feature extractors for this purpose. So, we use the VGG16 model \cite{simonyan2014very} to extract feature vectors from the masked input image and the generated image and then use these feature vectors to calculate the perceptual loss function of our model as:

\begin{equation}
L_{per} = E_{x,x'}[||F_P(x) - F_P(x')||_2] ,
\end{equation}

where the term $F_P(x)$ represents the feature vector of image $x$ extracted by the VGG16 model, $x$ represents the real image and $x'$ represents the output image of the generator model, and $L_{per}$ is defined as the average ($E_{x,x'}$) of the $L_2$ norm of the difference of the two feature vectors. \par

\textbf{De-identification loss:} This loss is used to prevent automatic face recognition systems from correctly matching the generated image with the true subject. To this purpose, we utilize a face recognition system to extract feature vectors from the real and synthesized images. The goal of the de-identification loss is then to maximize the distance between the extracted feature vectors of the input and output images, i.e. making the predicted identities as far as possible from each other. As a result, when a face recognition system tries to recognize the generated face image, it extracts features that are very far from the real image's features and cannot match the generated image to the input identity and incorrectly matches it to another person. In this paper, we use Resnet50 \cite{he2016deep} for face feature extraction. The de-identification loss function is defined as:

\begin{equation}
L_{did} = -E_{x,x'}[||F_{id}(x) - F_{id}(x')||_2]  
\end{equation}

where the term $F_{id}(x)$ represents the feature vector of image $x$ extracted by Resnet50, $x$ represents the real image and $x'$ represents the output image of the generator model. \par

\textbf{Final loss.} The final loss is defined as:
\begin{equation}
L_{final} = \lambda_{per}L_{per} + \lambda_{did}L_{did}
\end{equation}

where $\lambda_{per}$ and $\lambda_{did}$ are hyperparameters that control the relative importance of each loss. We have used $\lambda_{per} = \frac{1}{8800}$ and $\lambda_{did} = \frac{1}{12}$ in all of our experiments. By minimizing the final loss, firstly, the distance between the feature vectors extracted from the input image and the generated image by the VGG16 network \cite{simonyan2014very}  will be minimized, so the generated image will be perceptually similar to the input image. Secondly, the distance between the feature vectors extracted from the input image and the generated image by the Resnet50 network \cite{he2016deep} will be maximized, and hence the two images will be mapped to different identities. Finally, the reconstruction and de-identification tasks are performed simultaneously.

\subsection{Merge module}
The output image of the optimization module meets our goals but it lacks the background parts of the initial image that were removed in the face mask module. So, as the last step of our proposed model, we have devised a merge module to add the background from the original image to the de-identified face image generated by the optimization module to generate the final image. \par

The simplest way to merge the background and the face image is to perform a binary merge, taking the background pixels from the original image and the face pixels from the generated face image. As expected, this leads to a stark result as the output of the optimization module is not identical to its input and border of the face rectangle is detectable in the final image. On the other hand, applying smoothing filters to eliminate these unwanted edges will cause unpleasant blurring effects. \par

In this paper, we use a Fourier transform-based method \cite{szeliski2010computer} in the merge module shown in Figure \ref{block_diag} (c). In this method, the generated image and the input image are first convolved with a high-pass filter at different levels in Fourier domain, each with an increasing amount of sharpness. At the same time, the mask of the face part of the generated image is also convolved with a low-pass filter at different levels, each with an increasing amount of smoothness. Finally, at each level, high-pass filtered images are combined using a low-pass filtered mask as weights. The final result is obtained by adding the blended results at each level. The pseudocode of the merge algorithm is given in Algorithm \ref{mrg_alg}.\par
After getting the output image of the merge module, to have the best result, a histogram matching is performed so that the generated image's histogram matches the histogram of the input image \cite{gonzalez2018digital}. \par

\begin{algorithm}
\caption{The merge algorithm}
\label{mrg_alg}
\textbf{Input: }Input image A; generated image B.\par
\textbf{Input: }mask of the face part M.\par
\textbf{Input: }Gaussian filter function G.
\begin{algorithmic}[1]
\State Initialize $n = 10$, result matrix to zero.
\For{$l = 1$ to $n$}
    \State $G_A^l \gets G(A, l)$; \Comment{Building a Gaussian filtered version of the input image at level $l$}
    \State $G_B^l \gets G(B, l)$; \Comment{Building a Gaussian filtered version of the generated image at level $l$}
    \State $G_M^l \gets G(M, l)$; \Comment{Building a Gaussian filtered version of the mask image at level $l$}
    \If{$l = 1$}
    \State Continue;
    \EndIf
    \State $L_A^{l-1} = G_A^l - G_A^{l-1}$; \Comment{Building a Laplacian filtered version of the input image at level $l-1$}
    \State $L_B^{l-1} = G_B^l - G_B^{l-1}$; \Comment{Building a Laplacian filtered version of the generated image at level $l-1$}
    \State $C^{l-1} = G_M^{l-1} L_A^{l-1} + (1 - G_M^{l-1})L_B^{l-1}$
    \State $result = result + C^{l-1}$
\EndFor
\State \Return $result$
\end{algorithmic}
\end{algorithm}

\section{Experiments results and discussion}
\label{sec:experiments}
\subsection{Experimental setup}
\subsubsection{Datasets}
To evaluate the performance of de-identification methods, two face image datasets are used in this paper: Indian Celebrities dataset \cite{ghosh2020} and CelebA-HQ dataset \cite{karras2017progressive}. The details of the datasets are given below. \par

\textbf{Indian Celebrities dataset \cite{ghosh2020}:} 
This dataset contains 1487 images with 100 identities. This dataset is randomly divided into two parts in our experiments. The first part contains 442 images which cover all 100 identities. This part is considered for test and comparison of the models in the identity preservation task. The second part contains 1045 images which again include images of all 100 identities. This part is for training an SVM model for identification attack and comparing their identities with generated images for verification attack described in the following of this section. \par

\textbf{CelebA-HQ dataset \cite{karras2017progressive}:} The CelebA-HQ is the high-quality version of the CelebA dataset that consists of 30,000 images at 1024×1024 resolution. We only consider subjects with more than 12 images, resulting in 7743 images of 479 subjects. Similar to the previous dataset, we split this dataset into two parts: 2302 images for test and 5441 images for training the SVM model.
  
\subsubsection{Evaluation metrics} 
In this paper, we evaluate both the effectiveness of the models in concealing the identity of the face image and the quality of the generated images. To do so, we have employed three evaluation metrics defined below. \par

\textbf{The attack success rate (ASR):} This measure is defined as the fraction of the test images that are successfully de-identified by the model as follows:
  \begin{equation}
   ASR = \frac{No. \;\; of \;\; successfully\;\; de-identified\;\; images}{Total \;\; No. \;\; of\;\; test\;\; images }
   \end{equation}
The attack success rate is evaluated in two different scenarios in this paper. In the first scenario, called \emph{identification attack}, first, a feature extractor is used to extract feature vectors from train face images. Then, an SVM classifier, as in \cite{sun2018natural}, is trained with the extracted features to predict the identity of each input image. The di-identification model is then applied to the test images and the resulting images are fed into the face recognition SVM. The fraction of the misclassified images represents the success rate of the de-identification model. In the second scenario, called \emph{verification attack}, a match threshold $\tau$ is calculated based on the training data with false acceptance rate (FAR) of 0.1\% \cite{deb2020advfaces}. If the similarity (distance) of the feature vectors of two face images is more(less) than this threshold, the images are assumed to belong to the same person. Here, one image of the original images of a person is paired with a modified image of that person. The success rate is calculated as the number of images that are not matched with $\tau$ similarity (distance) threshold. \par

\textbf{Structural Similarity Index Measure (SSIM):} This metric is for measuring the similarity between two images computed as \cite{wang2004image},
   \begin{equation}
   SSIM(x, y) = \frac{(2\mu_{x}\mu_{y} + c_1)(2\sigma_{xy} + c_2)}{(\mu_x^2 + \mu_y^2 + c_1)(\sigma_x^2 + \sigma_y^2 + c_2)}
   \end{equation}
where $x$ and $y$ are two images, $\mu_x$ and $\mu_y$ are the means of images, $\sigma_x^2$ and $\sigma_y^2$ are the variances of the images, and $\sigma_{xy}$ is the covariance of $x$ and $y$. SSIM is a metric between -1 to 1 where -1 means the image pair are completely different and 1 signifies that the image pair are identical. Here, we employ SSIM to measure the similarity between the input and manipulated images. \par
   
\textbf{Peak Signal-to-Noise Ratio (PSNR):} PSNR is the ratio between the maximum possible power of an image and the power of corrupting noise that affects the quality of its representation. It is computed as \cite{hore2010image},
   \begin{equation}
   PSNR(x, y) = 10log_{10}(\frac{255^2}{MSE(x, y)})
   \end{equation}
where $x$ and $y$ are two images and $MSE(x,y)$ is the mean squared error (difference) of the two images. The higher values of the PSNR indicate higher image qualities and the smaller values imply large differences between to two images. Here, we use PSNR to compare the input and edited images.\par

\subsubsection{Face recognition systems} 
We report the result of the attack success rate for the identification and verification attacks on three different face recognition (FR) systems. The first utilized FR model is FaceNet \cite{schroff2015facenet} which gets a $160\times 160 \times 3$ input image and outputs an embedding with the length of 128. The second model is Resnet50 \cite{he2016deep} which gets a $224 \times 224 \times 3$ input image and outputs an embedding with the length of 2048. And finally, the third model is ArcFace \cite{deng2019arcface} that uses a similarity learning approach and angular margin loss to replace Softmax loss. ArcFace provides feature vectors with a length of 512. In our proposed model, ResNet50 FR model is treated as white-box model, and two other models are considered as black-box models as stated later in this section.

\subsubsection{GMFIM’s settings} For our experiments, StyleGAN2-encoder implementation \cite{rolux2019} is used as the face generator component of GMFIM. Before running the proposed model, first of all, images are aligned and the face parts are extracted from them. A gradient descent optimizer is used with a fixed learning rate of 1.0 in the optimization step. Each batch consists of one image, and in all experiments, we run the optimization module for 800 iterations. 

\subsection{Comparison with the state-of-the-art methods}
\subsubsection{Attack success rate} 
We use obfuscation model of AdvFaces \cite{deb2020advfaces} and Fawkes with high protection mode \cite{shan2020fawkes} as two state-of-the-art models to compare the results of GMFIM with the results of these two models. In Table \ref{table_de_identification_indian}, the attack success rates of the three models for the \emph{identification attack} scenario on the Indian Celebrities dataset are reported. In Table \ref{table_obfuscation_indian}, the results are reported for the \emph{verification attack} scenario on the Indian Celebrities dataset, with two different thresholds. Experiments performed on the Indian Celebrities dataset were repeated on the CelebA-HQ dataset and the results are shown in Table \ref{table_de_identification_celeb} and Table \ref{table_obfuscation_celeb}, respectively. As shown in these four tables, the behavior of the algorithms is almost the same for all test scenarios. Almost in all scenarios, GMIFM provides the best results and AdvFaces gives the lowest performance. The best results of GMIFM are obtained against Resnet50 which is obvious as it is a white-box model for GMIFM used in the optimization step of GMIFM. However, the results of the GMIFM against two other black-box FR models are close to the case of Resnet50 in most of the tests. Finally, GMFIM provides a higher attack success rate in all experiments compared to AdvFaces and Fawkes, except for the attacks against FaceNet where Fawkes provides better results. Albeit, Fawkes achieves this success rate by introducing extensive artifacts to the input image as discussed in the following. 

\begin{table}[htb]
\centering
\begin{tabular}{c c c c}
Model & FaceNet & Resnet50 & ArcFace \\
\hline\hline
AdvFaces & 28.5 & 31.9 & 40.45 \\
\hline
Fawkes & \textbf{88.23} & 80.09 & 67.19 \\
\hline
GMFIM & 70.13 & \textbf{91.4} & \textbf{71.94} \\
\hline
\end{tabular}
\caption{Attack success rate (\%) for identification attack on the Indian Celebrities dataset.}
\label{table_de_identification_indian}
\end{table}

\begin{table}[htb]
\centering
\begin{tabular}{c c c c}
Model & FaceNet & Resnet50 & ArcFace \\
\hline\hline
AdvFaces & 78.48 / 53.19 & 88.19 / 64.7 & 90.88 / 70.03 \\
\hline
Fawkes & \textbf{98.91 / 94.21} & 98.91 / 91.59 & 96.66 / 85.7 \\
\hline
GMFIM & 98.2 / 93.06 & \textbf{99.95 / 99.78} & \textbf{98.89 / 94.04} \\
\hline
\end{tabular}
\caption{Attack success rate (\%) for threshold computed at 0.1\% FAR / 1\% FAR for verification attack on the Indian Celebrities dataset.}
\label{table_obfuscation_indian}
\end{table}

\begin{table}[htb]
\centering
\begin{tabular}{c c c c}
Model & FaceNet & Resnet50 & ArcFace \\
\hline\hline
AdvFaces & 45.74 & 48.08 & 53.45 \\
\hline
Fawkes & \textbf{94.52} & 92.35 & 84.23 \\
\hline
GMFIM & 90.87 & \textbf{98} & \textbf{92.31} \\
\hline
\end{tabular}
\caption{Attack success rate (\%) for identification attack on the CelebA-HQ dataset.}
\label{table_de_identification_celeb}
\end{table}

\begin{table}[htb]
\centering
\begin{tabular}{c c c c}
Model & FaceNet & Resnet50 & ArcFace \\
\hline\hline
AdvFaces & 64.91 / 35.82 & 71.05 / 44.62 & 76.97 / 50.27 \\
\hline
Fawkes & \textbf{98.75 / 94.26} & 96.61 / 86.52 & 92.02 / 76.46 \\
\hline
GMFIM & 98.57 / 91.43 & \textbf{99.86 / 99.48} & \textbf{97.3 / 87.53} \\
\hline
\end{tabular}
\caption{Attack success rate (\%) for threshold computed at 0.1\% FAR / 1\% FAR for verification attack on the CelebA-HQ dataset.}
\label{table_obfuscation_celeb}
\end{table}

\begin{figure*}[t]
\centering
\includegraphics[width=0.7\textwidth,height=13cm]{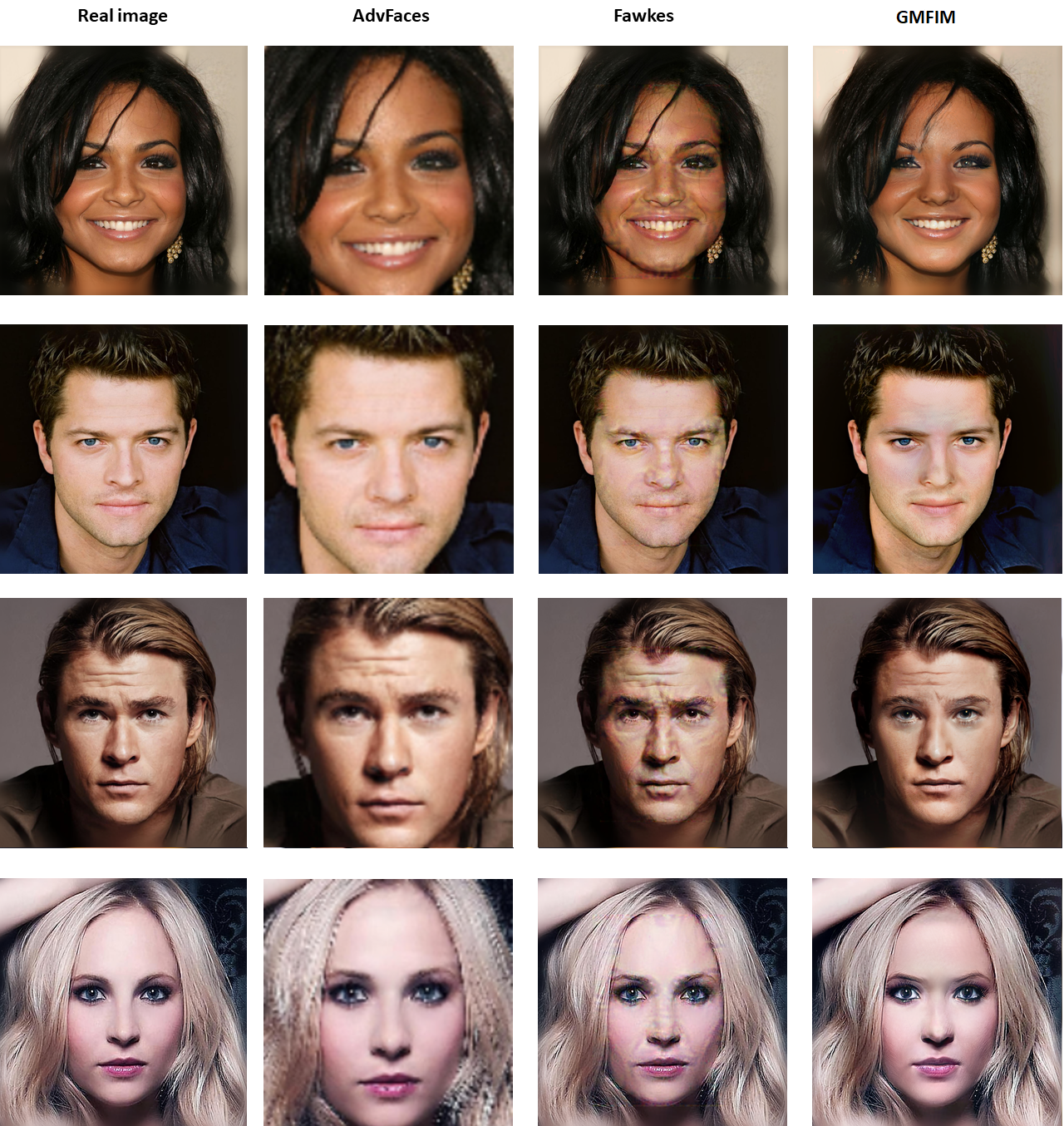}
\caption{Comparison of the quality of the generated images of the proposed GMFIM model and two other de-identification methods. From left to right, the first column shows the original image, and the next three columns show the results of AdvFaces \cite{deb2020advfaces}, Fawkes \cite{shan2020fawkes}, and GMFIM, respectively. All real images are selected from CelebA-HQ dataset \cite{karras2017progressive}.}
\label{compare_diag}
\end{figure*}

\begin{figure*}[t]
\centering
\includegraphics[width=0.6\textwidth,height=11cm]{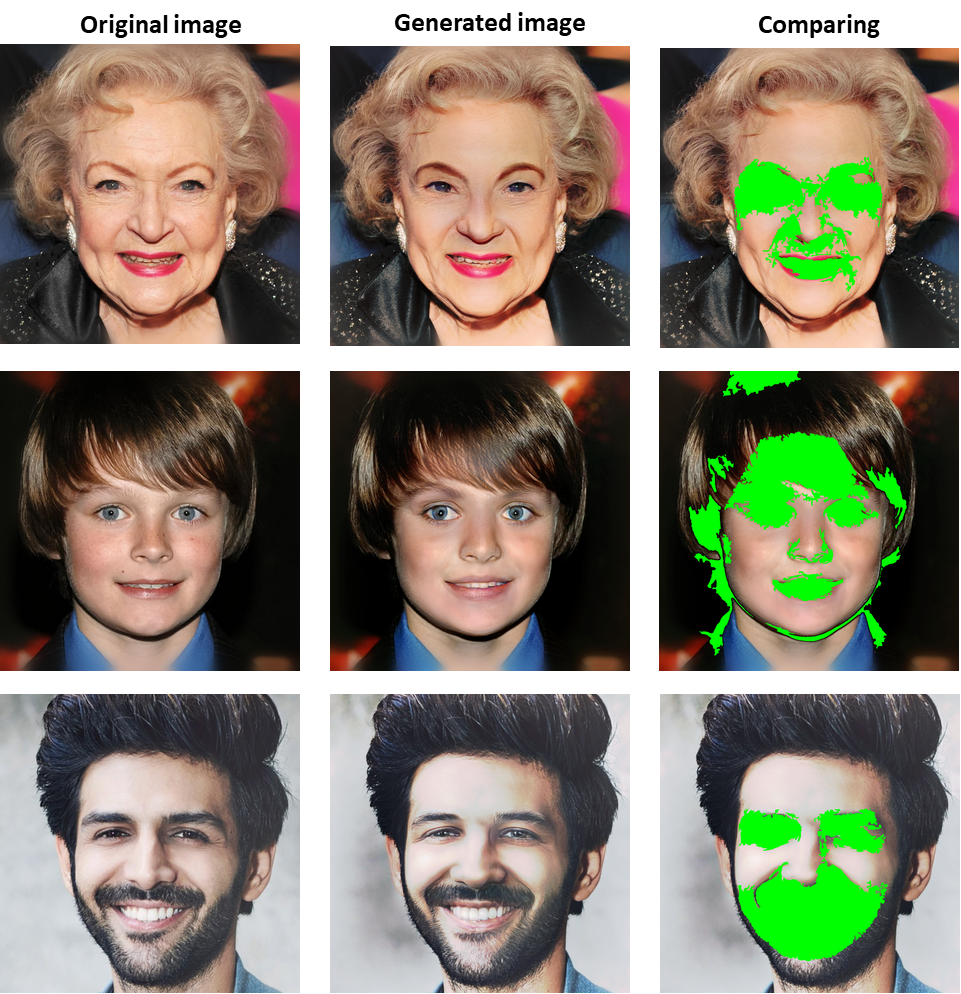}
\caption{Delineation of the changes made to the input images by GMFIM. The first column shows sample faces selected from CelebA-HQ dataset \cite{karras2017progressive} (the first two rows) and Celebrities dataset \cite{ghosh2020} (the third row). The second column shows represent the de-identified faces. In images of the third column represents the difference between the corresponding original and modified images. As it is clear from the results, the changes mainly occur in the main parts of the face, including the eyes, eyebrows, nose, and mouth. The changes include altering the color, size, or even the shape of the parts. The comparison of the original images with the generated images shows that instead of drastically modifying a single part or a small region of the face, most parts of the faces are slightly modified by the proposed model.}
\label{analysis_change}
\end{figure*}

\begin{figure*}[t]
\centering
\includegraphics[width=0.6\textwidth,height=11cm]{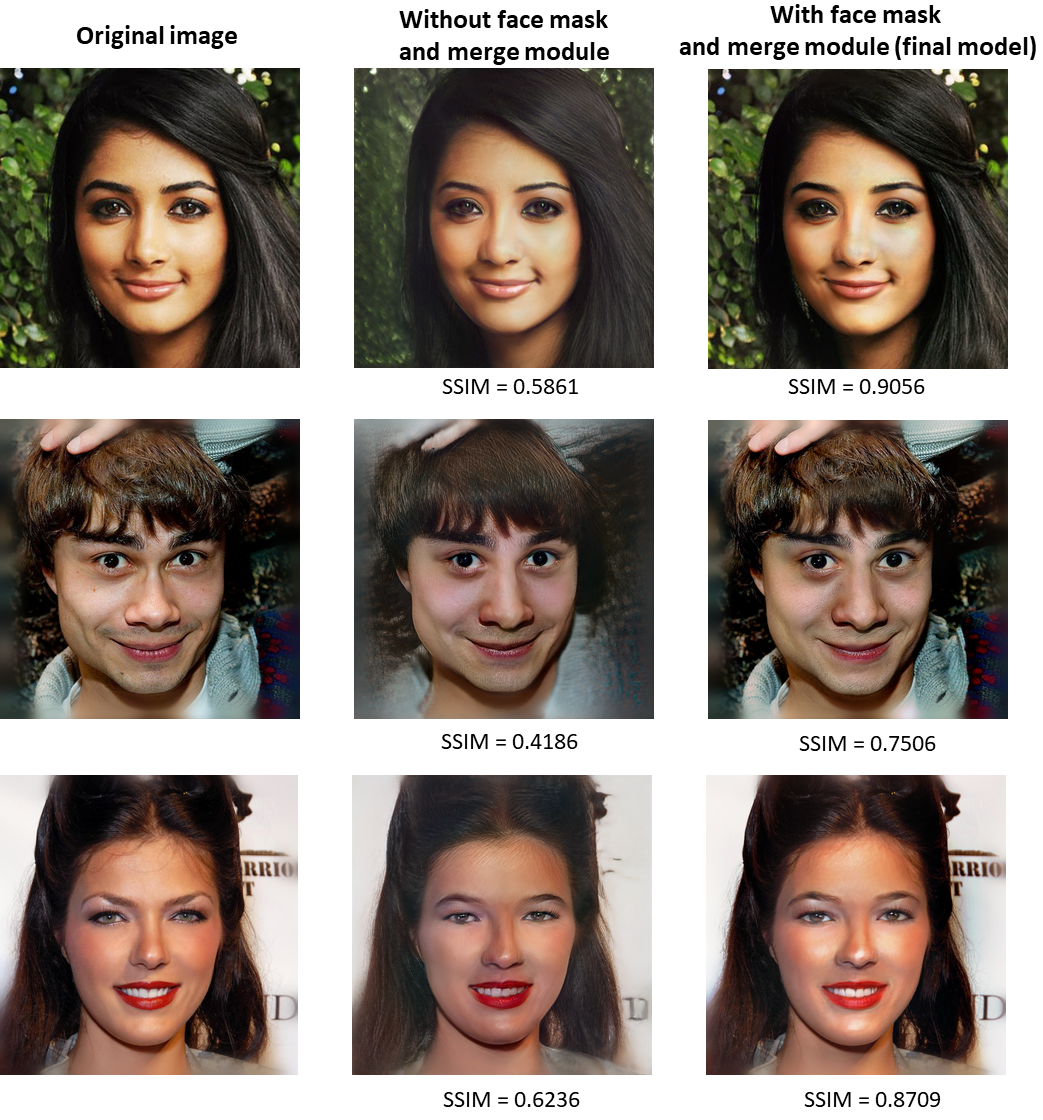}
\caption{The effect of the face mask and the merge modules of GMFIM on the generated image. The left column show the original images from Indian Celebrities dataset \cite{ghosh2020} (the first row) and CelebA-HQ dataset \cite{karras2017progressive} (the second and third rows). The middle and right columns respectively show the results of the GMFIM without the face mask and merge modules and the full GMFIM model. The SSIM of each modified image is also reported below the image. Without using the face mask module and the merge module, the whole input image has to be reconstructed by the optimization module. This causes the generated image to have a blurry background, a smaller SSIM, and in some cases less quality in the face part.}
\label{analysis_effect_modules}
\end{figure*}

\begin{figure*}[t]
\begin{center}
\includegraphics[width=0.5\textwidth,height=6cm]{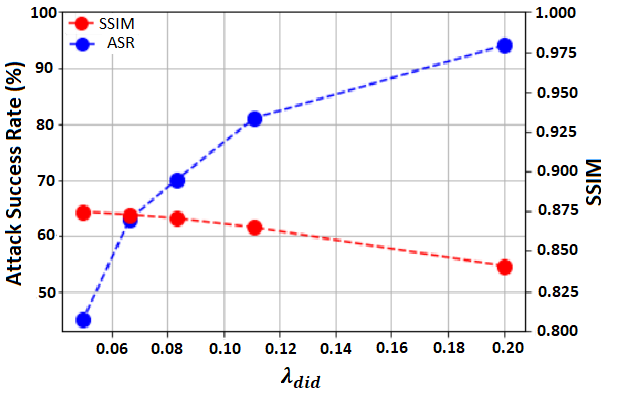}
\caption{The effect of $\lambda_{did}$ on SSIM and attack success rate (ASR) of GMFIM for $\lambda_{per}=\frac{1}{8800}$. Increasing the value of $\lambda_{did}$ degrades the quality of the manipulated images and reduces the SSIM but increases the attack success rate. This is because the importance of the de-identification loss increases in the final loss. For all experiments $\lambda_{did}=\frac{1}{12}\approx0.083$ has been used.}
\label{landa_effect}
\end{center}
\end{figure*}

\subsubsection{Image quality} For comparing the quality of generated images from our model with the state-of-the-art models, SSIM and PSNR metrics have been measured from the original and manipulated images for all models. The results are summarized in Table \ref{image_quality}. As the results show, the SSIM and PSNR of GMFIM are close to but less than those of the two other models. However, this doesn't mean that our images have less quality.  Figure \ref{compare_diag} shows samples of the generated images by GMFIM, AdvFaces and Fawkes. As it can be seen in this figure, generated images of GMFIM have the best quality and are more appealing to the human eye. It is clear from these few samples that the face images generated by AdvFaces are generally blurred and the results of Fawkes are full of annoying artifacts.\par

As stated before, AdvFaces creates adversarial masks that are added to the original image. Fawkes also creates perturbations that are added to the face image. So, both models are trying to create a mask and add it to the face for identity hiding. On the contrary, GMFIM tries to find a latent vector that generates the face image with the desired attributes. So, compared to the original image, our generated image may have large changes in small regions or small changes in large areas of the face which are human-imperceptible but fool the face recognition system. The reason for less SSIM and PSNR in our results is these changes.  These changes are discussed in more detail in the rest of this section.\par

\begin{table}[htb]
\centering
\begin{tabular}{c c c }
Model & SSIM & PSNR \\
\hline\hline
AdvFaces & 0.9452  & 33.36   \\
\hline
Fawkes & 0.9853  & 37.23  \\
\hline
GMFIM & 0.8699  & 29.91  \\
\hline
\end{tabular}
\caption{Image quality for different models on the Indian Celebrities dataset.}
\label{image_quality}
\end{table}

\subsection{Analysis}
\textbf{Comparison of the manipulated and the original images:} As stated before, the GMFIM model is not based on additive masks; instead, it tries to find a latent vector to generates a face image with the desired attributes. As a result of this process, the regenerated face image is different from the original one in different regions. To illustrate these changes, a comparison between some input images and their corresponding manipulated versions by GMFIM are shown in Figure \ref{analysis_change}. As shown in this figure, the changes mainly occur in the main parts of the face including the eyes, eyebrows, nose, and mouth. The amount and severity of the modification are adaptively determined for each image by the optimization process. The differences of the images include changes in the color, size, and even the shape of facial parts. The main point is that most of these changes are not tangible for human users of the social media specially when the original images are not present for comparisons and they easily identify the manipulated images.

\textbf{The importance of the face mask and merge modules:} As mentioned earlier, the goal of the face mask module is to help the optimization module to focus only on the face part of the input image. This in turn necessitates the use of a merge module to add the separated background from the original image to the de-identified face image. Without this separation, the GAN-based optimization step has to reconstruct the whole input image which makes the generator much more complicated. For example, while the StyleGAN is very successful in regenerating diverse face photos, it fails to properly reconstruct many background patterns. Three examples of the results of the two aforementioned approaches are shown in Figure \ref{analysis_effect_modules}. As shown in this figure, omitting the mask-guidance part of the GMFIM causes a quite blurry background with a much smaller SSIM (right images) compared to the complete model (middle images). Moreover, trying to reconstruct the whole image in the optimization module reduces the quality of the generated face regions in this case. \par

\textbf{The effect of the hyperparameters:} $\lambda_{per}$ and $\lambda_{did}$ are two main hyperparameters of GMFIM that control the relative importance of the perceptual and de-identification losses, respectively. With the help of these two hyperparameters, the quality of the generated images and the success rate against recognition systems can be controlled. \par
To observe the effect of changing these hyperparameters, the value of $\lambda{per}$ is fixed at $\frac{1}{8800}$ and different values are assigned to $\lambda_{did}$. The SSIM and the attack success rate (ASR) of GMFIM on Indian Celebrities dataset for different values of  $\lambda_{did}$  are reported in Figure \ref{landa_effect}. As the results show, increasing the value of $\lambda_{did}$ degrades the quality of the manipulated images and results in lower SSIMs. Also, it increases the probability of successfully hiding the identity of the person and higher attack success rates. The reason for this is that by increasing the value of $\lambda_{did}$, the importance of de-identification loss increases and it plays a more important role in generating the final image. On the contrary, decreasing the value of $\lambda_{did}$ improves the quality of the generated images but the produced face image is more likely to be identified by AFR systems. As mentioned earlier, in the experiments of this paper, $\lambda_{did}=\frac{1}{12}\approx0.083$ has been used. \par

\section{Conclusion}
In this paper, we proposed a new model called GMFIM for hiding the identity of face images to help protecting the privacy of the people on social media applications. We tried to make the output images of our model as similar as possible to the input images so that the manipulated images can be still shared on the social nets. The experimental results show that GMFIM can reach high attack success rates without degrading the quality of the input image. However, as stated before, the face images are generated by a pre-trained GAN model and the quality of the generated images and the diversity of the face images supported by the model are limited by this face generator component. Albeit, it should be noted that the proposed model is not restricted to a special face generator and it is possible to adopt the state-of-the-art face generator in proposed model. Finally, it should be noted that the model does not guaranty to conceal the identity from all face recognition methods and there is still space for improvement. 

\bibliographystyle{unsrt} 
\bibliography{references}

\end{document}